\documentclass[runningheads]{llncs}
\usepackage[T1]{fontenc}
\usepackage{graphicx}
\usepackage{hyperref}
\usepackage{multirow}
\usepackage{amsmath}
\usepackage{array,booktabs,adjustbox} 
\usepackage{url}
\newcolumntype{C}[1]{>{\centering\arraybackslash}p{#1}}

\newcommand{\imgw}{2.7cm}   
\newcommand{\etal}{\textit{et al}}

\newcommand{\imgcell}[1]{%
  \fbox{\includegraphics[width=\imgw,keepaspectratio]{#1}}%
}

\title{ICPR 2026 Competition on Privacy-Preserving Person Re-Identification from Top-View RGB-Depth Camera (TVRID)\thanks{Author preprint. The final authenticated version will appear in the ICPR 2026 proceedings published by Springer.}}
\titlerunning{ICPR 2026 TVRID}

\author{Raphaël Delécluse\inst{1,2}\and
Hazem Wannous\inst{1}\and
Laurent Guimas\inst{2}}
\authorrunning{R. Delécluse et al.}
\institute{IMT Nord Europe, University of Lille, CNRS UMR 9189 - CRIStAL, F-59000 Lille, France \and
Explain, F-59000 Lille, France}

\begin{document}
\maketitle



\begin{abstract}
This companion paper reports the ICPR 2026 TVRID competition on privacy-aware top-view person re-identification. We present the competition setting, the released RGB-Depth dataset, and a summary of final results with descriptions of the top entries.

TVRID contains 86 identities captured by four synchronized overhead Intel RealSense D455 cameras, with paired RGB/Depth streams and structured geometric variation across flat, ascent, descent, and oblique viewpoints. The evaluation protocol includes three tracks: RGB Re-ID, Depth Re-ID, and RGB$\leftrightarrow$Depth cross-modal retrieval. Submissions are ranked using mAP and CMC-1 under a unified server-side evaluation.

The final results show a clear difficulty ordering (RGB $>$ Depth $>$ Cross-Modal), highlighting both the challenge of modality-constrained retrieval and the feasibility of strong performance with modality-invariant learning. By releasing the dataset at \url{https://zenodo.org/records/17909410}, the evaluation scripts at \url{https://github.com/RaphaelDel/ICPR-TVRID}, and the accompanying documentation, TVRID establishes a reproducible benchmark for top-view, depth-based, and cross-modal person re-id.
\end{abstract}

\section{Introduction}
Person re-identification (Re-ID) studies how to determine whether observations from different, non-overlapping cameras correspond to the same individual. It is a central problem in modern computer vision, with direct impact on surveillance, security, mobility analytics, and crowd understanding \cite{gong_person_2011,zheng_scalable_2015,ye_deep_2022,leng_survey_2020}. Over the last decade, progress has been substantial, especially with deep metric learning, part-based modeling, and transformer architectures that learn discriminative identity embeddings from RGB data. Yet, this progress is still strongly tied to side-view visual assumptions, where clothing color and texture carry much of the identity signal.

This dependency creates two limitations. First, RGB-based Re-ID remains vulnerable to difficult real-world factors such as strong viewpoint changes, short observations, and occlusion \cite{ahmed_improved_2015,rao_hierarchical_2024}. Second, RGB sensing raises persistent privacy concerns in public spaces because it can expose personally identifiable appearance cues. These issues are particularly acute in overhead deployments, where camera placement is often selected precisely to reduce facial exposure while still enabling operational monitoring.

Depth sensing is therefore a compelling direction for privacy-aware Re-ID. Compared with RGB imagery, depth captures 3D structure and body geometry, is less sensitive to illumination changes, and naturally suppresses color-based identity cues \cite{wu_robust_2017,jia_2d_2022,wu_end--end_2022}. In top-view configurations, this modality is especially relevant because the available information is already dominated by anthropometric shape, posture, and motion patterns rather than frontal appearance \cite{delecluse_privacy-preserving_2025,lejbolle_multimodal_2017,paolanti_person_2018,liciotti_person_2017}. In other words, top-view depth Re-ID reframes the problem: it is no longer mainly about matching clothes, but about learning robust geometry-aware and motion-aware identity representations.

Despite this motivation, dataset support for top-view depth-aware Re-ID remains limited. Public resources such as TVPR2 \cite{martini_open-world_2020}, GODPR \cite{fuentes-jimenez_depth_2020}, and BIWI RGBD-ID \cite{munaro_one-shot_2014} have enabled important methodological advances, but they do not provide a unified protocol explicitly centered on controlled elevation transitions. This point matters because ascent and descent introduce systematic geometric shifts that are common in practice (stairs, ramps, level changes) and can significantly alter perceived body proportions and trajectory dynamics.

The ICPR 2026 TVRID challenge was designed to address this specific gap. TVRID provides synchronized RGB and Depth streams acquired by four overhead cameras spanning flat, ascent, descent, and oblique viewpoints, with consistent identity annotations and IN/OUT observations. The challenge evaluates three complementary tasks: RGB Re-ID, Depth Re-ID, and RGB$\leftrightarrow$Depth cross-modal retrieval. This design makes it possible to study not only raw recognition performance, but also privacy--performance trade-offs and modality-invariant generalization under realistic geometric variation.

This paper reports the finalized outcomes of the competition and presents TVRID as a reproducible benchmark for the community. Beyond presenting rankings, we analyze how performance changes across modalities and camera geometry, and we discuss what these findings imply for privacy-aware deployment of person Re-ID systems in real environments.

\section{Competition overview}

\subsection{Dataset Description}
\label{sec:dataset}

TVRID is designed to address limitations of existing top-view Re-ID datasets, which typically emphasize flat-ground trajectories with limited geometric variation. In contrast, TVRID explicitly introduces structured viewpoint and elevation changes in a synchronized RGB--Depth setting, making it suitable for analyzing both privacy-preserving and cross-modal retrieval behavior.
The dataset contains 86 identities recorded by four overhead Intel RealSense D455 cameras at 15 FPS, with synchronized RGB and Depth streams. For each passage, subjects are captured twice per camera (IN/OUT), producing eight observation sets per passage (four cameras $\times$ IN/OUT). A typical passage lasts about \(\sim\)15\,s, with approximately 2--3\,s of effective field-of-view per camera.
Compared with prior top-view benchmarks, TVRID contributes: (i) structured elevation variation (flat $\rightarrow$ ascent $\rightarrow$ descent $\rightarrow$ oblique), (ii) consistent identity annotations across all cameras and directions, and (iii) paired synchronized RGB/Depth releases supporting both modality-specific and cross-modal protocols.

\begin{figure}[t]
    \centering
    \includegraphics[width=0.85\textwidth]{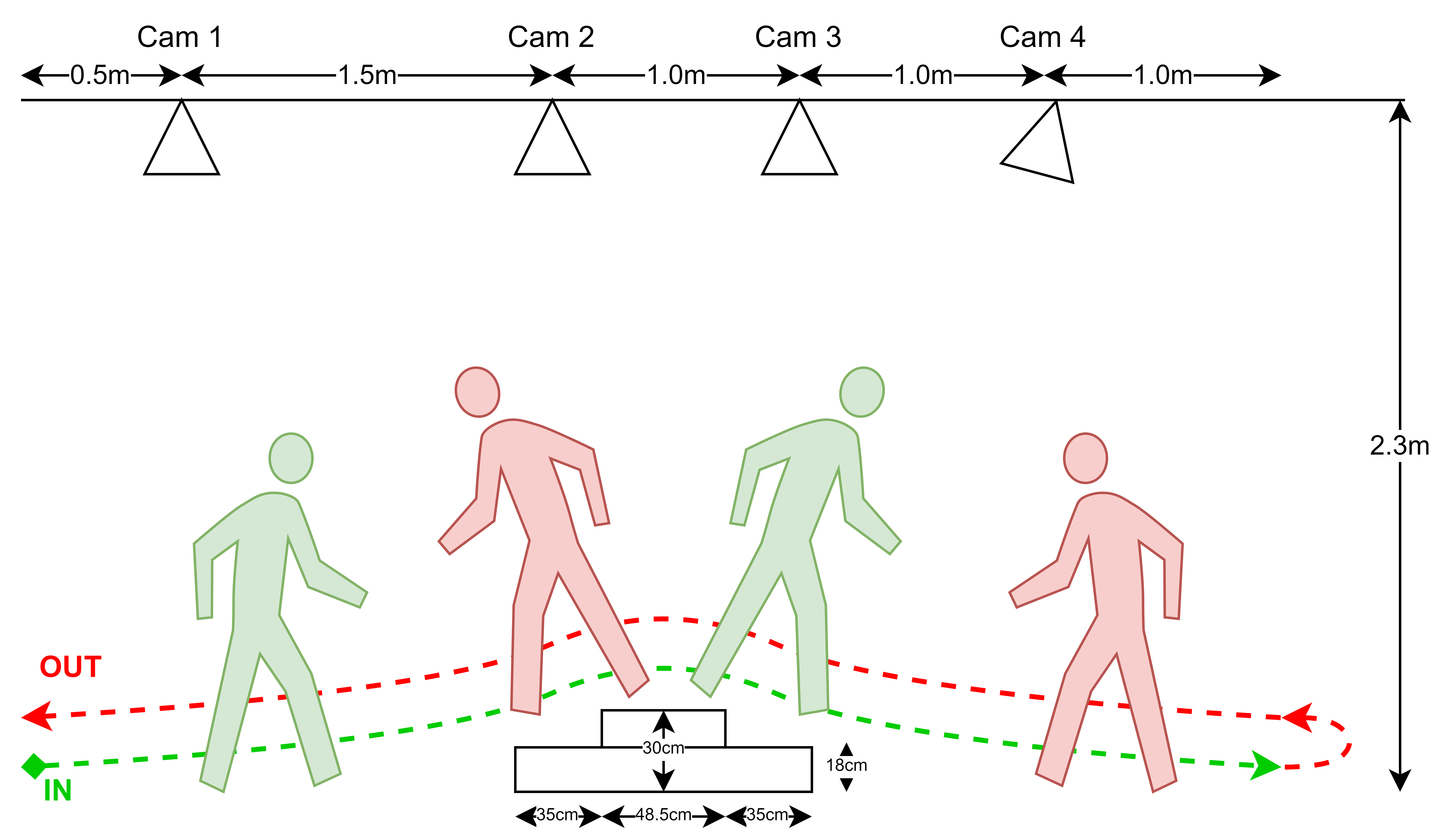}
    \caption{Acquisition schematic. Top-view layout with four overhead cameras along a short path including a stepladder segment. Distances are chosen to obtain successive, partially overlapping observations with distinct geometry (flat, ascent, descent, oblique).}
    \label{fig:hirid_setup}
\end{figure}

\paragraph{Acquisition geometry and camera roles.}
Cameras are mounted top-view along a short path including a stepladder segment, as it can be seen in Figure~\ref{fig:hirid_setup}. Distances provide partially overlapping windows while inducing distinct geometry across cameras:

\begin{enumerate}
    \item Camera~1 (flat ground): canonical level-ground passage;
    \item Camera~2 (ascent):  upward motion on a stepladder.
    \item Camera~3 (descent): downward motion on the stepladder.
    \item Camera~4 (oblique roof view): elevated oblique viewpoint.
\end{enumerate}

\begin{figure}[t]
\centering
\setlength{\tabcolsep}{4pt}
\renewcommand{\arraystretch}{1.05}

\newcommand{\cambox}[4]{%
  \begin{tabular}{C{.48\linewidth} C{.48\linewidth}}
    \textbf{IN} & \textbf{OUT} \\
    \imgcell{#1} & \imgcell{#2} \\
    \imgcell{#3} & \imgcell{#4} \\
  \end{tabular}%
}

\begin{minipage}[t]{0.48\linewidth}\centering
\textbf{Camera 1}\\[2pt]
\cambox
{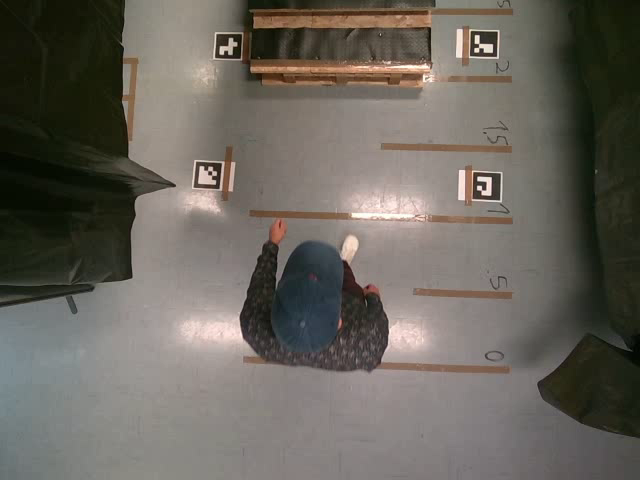}
{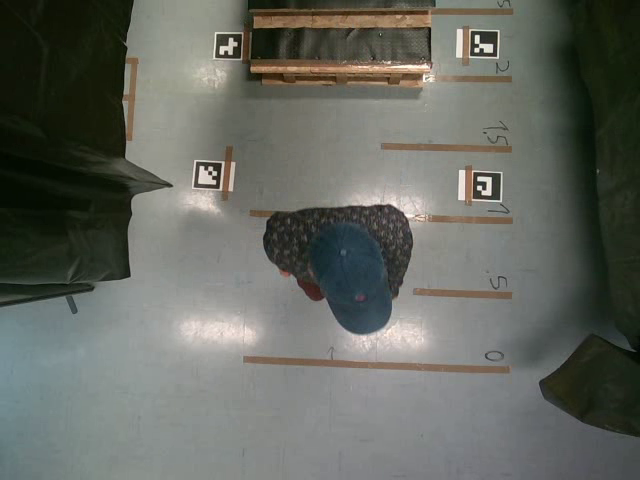}
{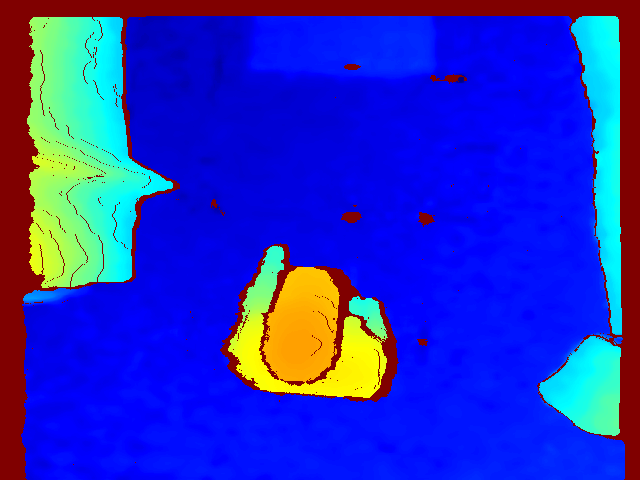}
{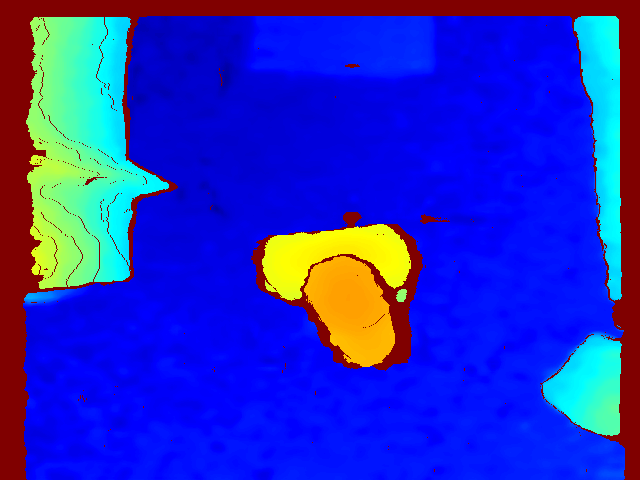}
\end{minipage}\hspace{0.02\linewidth}
\begin{minipage}[t]{0.48\linewidth}\centering
\textbf{Camera 2}\\[2pt]
\cambox
{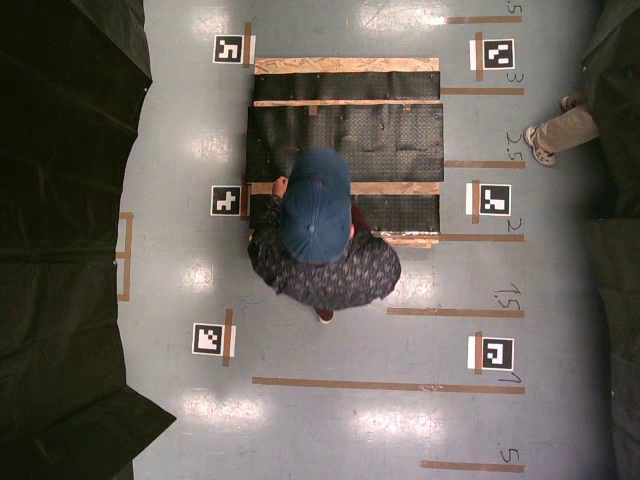}
{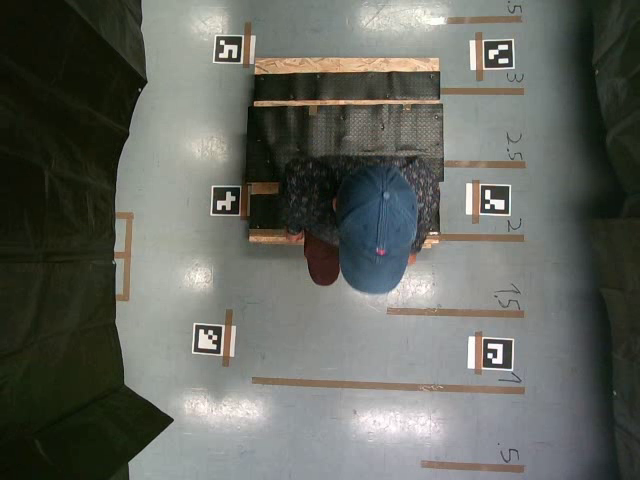}
{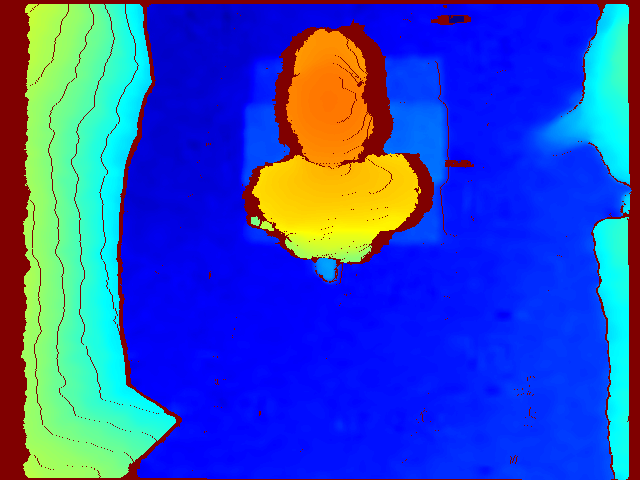}
{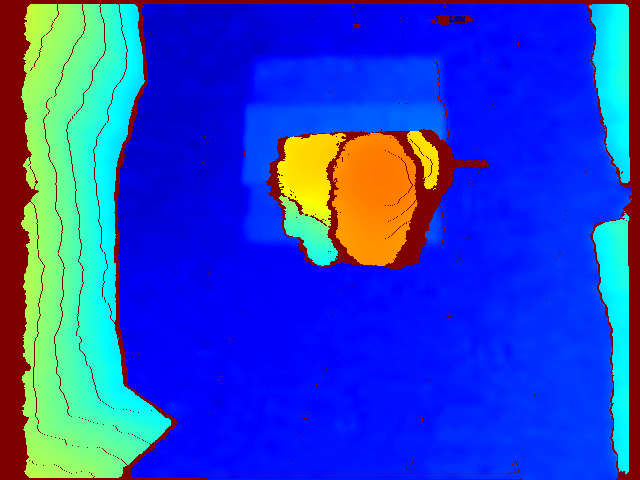}
\end{minipage}

\vspace{6pt}

\begin{minipage}[t]{0.48\linewidth}\centering
\textbf{Camera 3}\\[2pt]
\cambox
{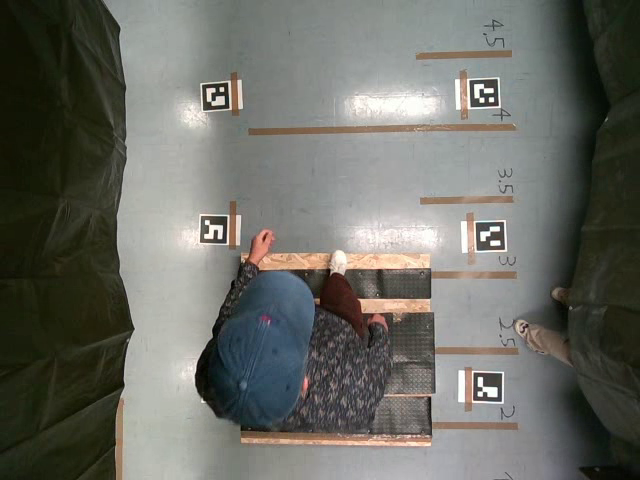}
{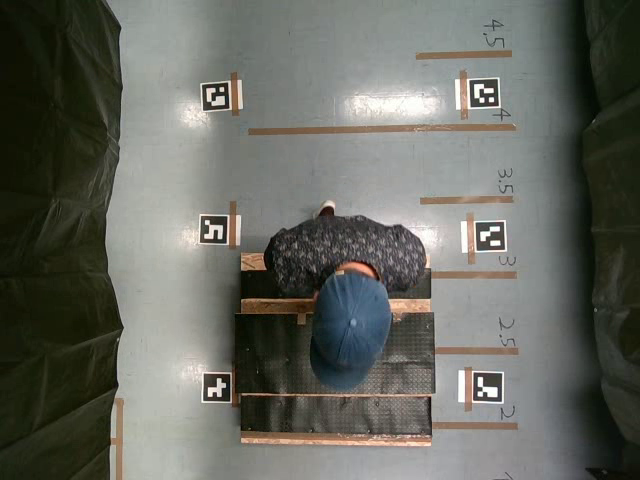}
{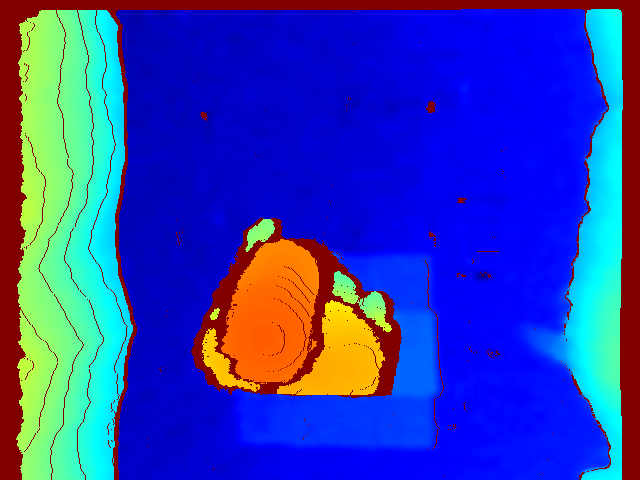}
{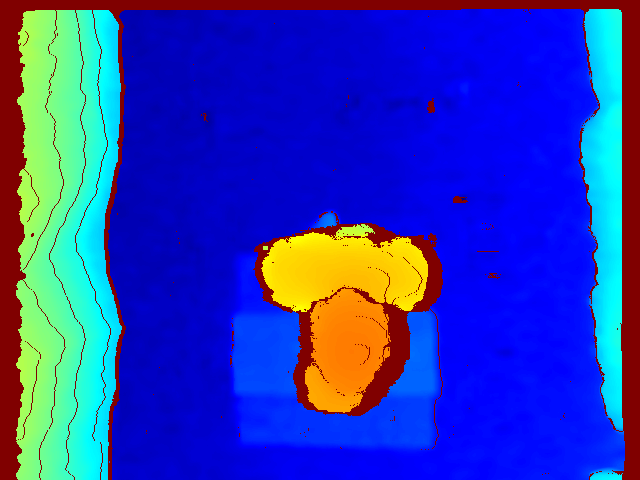}
\end{minipage}\hspace{0.02\linewidth}
\begin{minipage}[t]{0.48\linewidth}\centering
\textbf{Camera 4}\\[2pt]
\cambox
{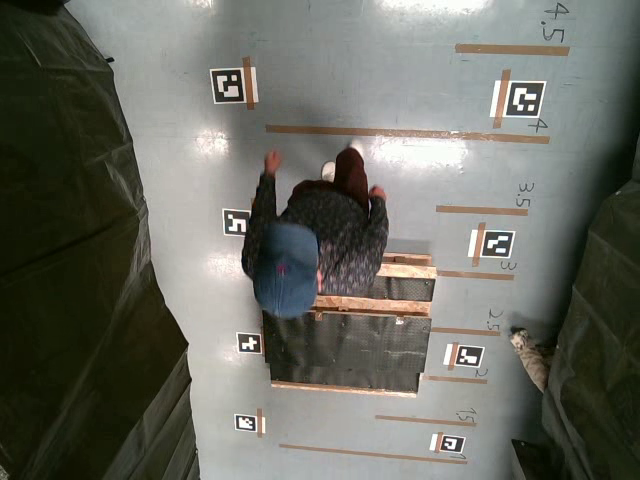}
{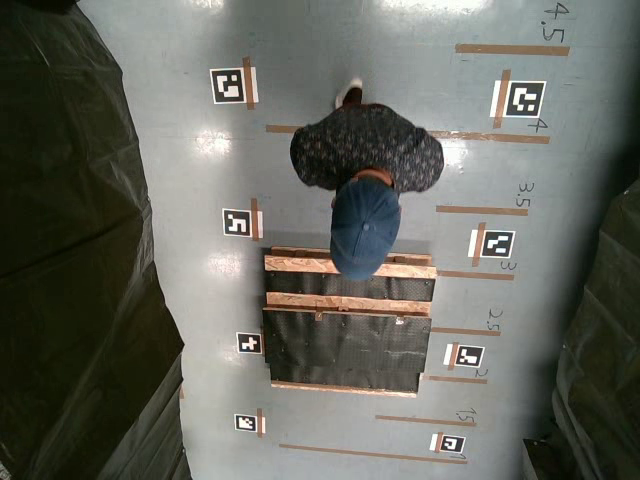}
{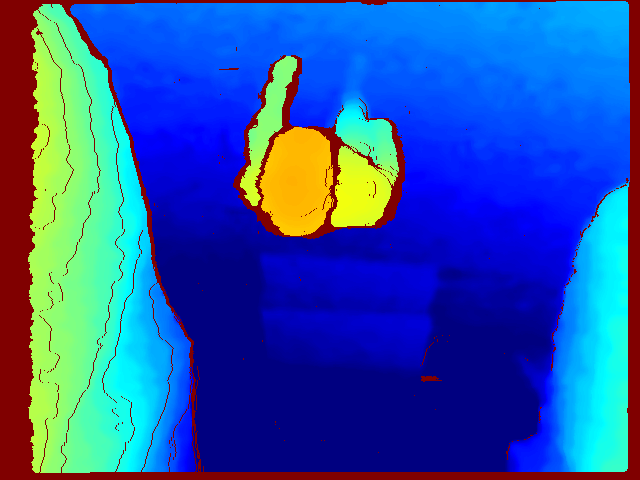}
{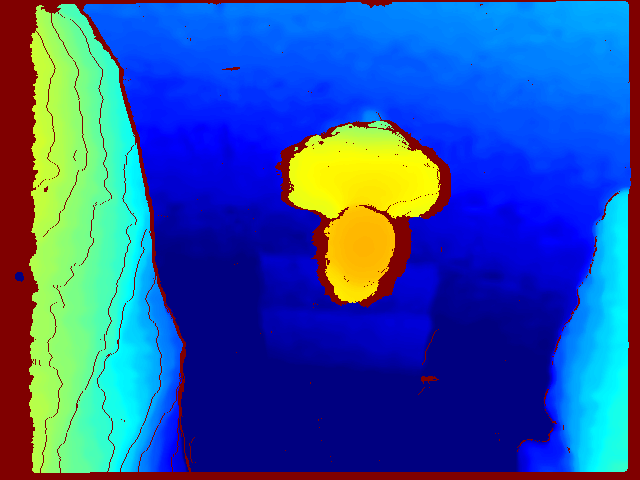}
\end{minipage}

\caption{2×2 camera layout. Each camera: RGB row (top) and Depth row (bottom), with IN/OUT columns.}
\label{fig:hirid_grid}
\end{figure}

\noindent RGB/Depth frames for each camera and direction are shown in Fig.~\ref{fig:hirid_grid}. The IN/OUT pairing per camera is used explicitly in the retrieval protocols described in Section~\ref{sec:evaluation}.


\paragraph{Modalities, synchronization and releases.}
RGB and Depth streams are recorded synchronously; all frames are timestamped and aligned at acquisition. Alongside raw streams (640$\times$480), a depth-driven detector provides person crops to remove the need for a separate detection stage; raw data remain available for end-to-end pipelines.




\subsection{Description of Tracks}
\label{sec:tracks_description}
Three official tracks were evaluated:
\begin{enumerate}
    \item \textbf{Track 1 -- RGB Re-ID:} Standard person re-identification from RGB images.
    \item \textbf{Track 2 -- Depth Re-ID:} Privacy-preserving re-identification from depth maps.
    \item \textbf{Track 3 -- RGB$\leftrightarrow$Depth Cross-Modal Re-ID:} Query in one modality (RGB) and retrieval in the other modality (Depth), testing modality-invariant representations.
\end{enumerate}

Track~1 -- RGB: is included as a comparative reference condition. Strong performance is already expected for RGB-based Re-ID in many settings, so this track provides an upper-bound-like baseline for interpreting the two more challenging settings.

Track~2 -- Depth: addresses the central scientific question of the competition: can a Re-ID system operate accurately using \emph{depth only}, under realistic geometric variation? This question is particularly relevant in TVRID because elevation changes and viewpoint shifts alter apparent body geometry, and height-related cues could dominate depth perception in a way that may not generalize across cameras. The track therefore evaluates whether modern representations can extract sufficiently discriminative identity information from depth maps beyond simple height cues.

Track~3 -- RGB$\leftrightarrow$Depth cross-modal: is intentionally formulated as a hard task. It tests whether identity information can be aligned across heterogeneous modalities, where query and gallery distributions differ by design. In practice, this track probes modality-invariant representation learning under strong domain mismatch and serves as a stress test for methods targeting robust deployment across sensing configurations.

\subsection{Evaluation Metrics}
\label{sec:evaluation}



Cumulative Matching Characteristic (CMC) at Rank-1 and the mean Average Precision (mAP) was selected as evaluation metric in this competition. These are the prevailing metrics in state-of-the-art top-view Re-ID benchmarks\cite{paolanti_deep_2020,lejbolle_person_2020,martini_open-world_2020,mukhtar_cmot_2024,delecluse_privacy-preserving_2025}. Theses metrics are described as follows:

\medskip
\noindent\textbf{CMC@k (Rank-k).}
Let $\mathcal{Q}$ be the set of queries and, for a query $q\in\mathcal{Q}$, let $\mathcal{G}_q\subseteq\mathcal{G}$ denote the set of gallery images sharing the same identity as $q$ (excluding duplicates). Let $r_q$ be the rank position of the first correct match for $q$ in the gallery ranking. The CMC at cutoff $k$ is
\[
\mathrm{CMC@}k \;=\; \frac{1}{|\mathcal{Q}|}\sum_{q\in\mathcal{Q}} \mathbf{1}\!\left[r_q \le k\right].
\]
Intuitively, CMC@1 is the fraction of queries whose best match is retrieved at rank 1.

\medskip
\noindent\textbf{Mean Average Precision (mAP).}
For a query $q$, consider the ranked gallery list $\{g_1,\dots,g_N\}$. Define $\mathrm{rel}_q(k)=1$ if $g_k\in\mathcal{G}_q$ and $0$ otherwise, and let $P_q(k)$ be the precision at cutoff $k$:
\[
P_q(k) \;=\; \frac{\sum_{i=1}^{k} \mathrm{rel}_q(i)}{k}.
\]
The Average Precision (AP) for $q$ is
\[
\mathrm{AP}(q) \;=\; \frac{1}{|\mathcal{G}_q|} \sum_{k=1}^{N} P_q(k)\,\mathrm{rel}_q(k),
\]
and the mean Average Precision is
\[
\mathrm{mAP} \;=\; \frac{1}{|\mathcal{Q}|}\sum_{q\in\mathcal{Q}} \mathrm{AP}(q).
\]

Because the hidden test subset is relatively limited in size, evaluation is organized into three complementary camera-based scenarios. (i) Within-camera IN/OUT: queries from the IN direction are matched against OUT samples from the same camera (and vice versa). (ii) Ascent vs. descent: samples from the ascent viewpoint are matched against samples from the descent viewpoint to explicitly test robustness to elevation-related geometry changes. (iii) Flat vs. oblique: samples from the flat camera are matched against the oblique camera to assess viewpoint generalization.

For each scenario, we compute CMC@1 and mAP, and the final score for a track is obtained by averaging the scenario-level CMC@1 values and mAP values. Duplicate frames of the same identity/instant are excluded from valid query--gallery matches.
The official evaluation script is available on GitHub at \url{https://github.com/RaphaelDel/ICPR-TVRID}.


\subsection{Submission system}

Codabench\cite{xu2022codabench} was used as the submission platform. Codabench is an open-source system that efficiently supports the organization of scientific competitions. Participants could test their methods and receive real-time feedback through a competitive leaderboard. Each track had its own leaderboard metrics, and participants also had access to more detailed metrics and diagnostic plots.

The competition was organized in two distinct phases: a two-month development phase, during which models were evaluated on an unseen validation set with a quota of five submissions per team per day, followed by a one-week final test phase, in which each team could submit only one selected approach for evaluation on the unseen test set.

\section{Competition Outcomes}

\begin{table}[t]
\centering
\caption{Participant-level results across the three tracks; participant names are reported using their CodaBench submission usernames.}
\label{tab:participant_full_results}
\scriptsize
\setlength{\tabcolsep}{4pt}
\resizebox{\textwidth}{!}{%
\begin{tabular}{@{}lcccccc@{}}
\toprule
\multirow{2}{*}{\textbf{Participant}} & \multicolumn{2}{c}{\textbf{RGB Track}} & \multicolumn{2}{c}{\textbf{Depth Track}} & \multicolumn{2}{c}{\textbf{Cross-Modal Track}} \\
\cmidrule(lr){2-3}\cmidrule(lr){4-5}\cmidrule(lr){6-7}
& \textbf{CMC@1} & \textbf{mAP} & \textbf{CMC@1} & \textbf{mAP} & \textbf{CMC@1} & \textbf{mAP} \\
\midrule
jiangvicky & 99.67\% & 99.84\% & 98.85\% & 99.37\% & 97.89\% & 98.76\% \\
lavia\_lab12 & 96.74\% & 98.15\% & -- & -- & -- & -- \\
rashid & 99.17\% & 99.51\% & 90.44\% & 94.33\% & 99.84\% & 99.92\% \\
es\_code26 & 99.84\% & 99.92\% & 97.19\% & 98.34\% & 91.50\% & 94.88\% \\
abhinavkumar03 & 99.84\% & 99.88\% & 90.42\% & 93.51\% & -- & -- \\
hiensumi & 100.00\% & 100.00\% & 81.84\% & 88.83\% & -- & -- \\
sjkdgfhl\_ & 99.35\% & 99.63\% & 80.34\% & 86.26\% & -- & -- \\
liuyiping\_mia & 93.23\% & 94.01\% & 94.00\% & 93.47\% & 93.36\% & 93.69\% \\
baondbk & 94.56\% & 92.31\% & -- & -- & -- & -- \\
liangying & 98.85\% & 99.37\% & 86.62\% & 91.77\% & 75.82\% & 84.66\% \\
1105linyufan & 98.85\% & 99.37\% & 82.99\% & 89.22\% & 72.18\% & 82.47\% \\
yufanlin & 97.71\% & 98.58\% & 84.39\% & 90.06\% & 67.89\% & 78.95\% \\
trantruongmmcii & 99.51\% & 99.75\% & 76.00\% & 84.10\% & 19.70\% & 37.10\% \\
jrguo & 96.36\% & 97.50\% & 45.60\% & 60.78\% & 10.53\% & 27.51\% \\
carlometta & 100.00\% & 99.98\% & 47.28\% & 61.64\% & 7.06\% & 22.12\% \\
se192003 & 72.22\% & 81.51\% & 30.55\% & 49.83\% & 21.33\% & 38.85\% \\
sonu6799 & 90.07\% & 92.88\% & 22.41\% & 37.37\% & 11.22\% & 24.50\% \\
vivek\_katuri & 77.67\% & 85.20\% & -- & -- & 4.33\% & 15.53\% \\
chari7890 & 90.07\% & 92.88\% & 12.68\% & 28.67\% & 11.22\% & 24.50\% \\
prajyoth4567 & 90.07\% & 92.88\% & 22.41\% & 37.37\% & 4.33\% & 15.53\% \\
aruna5678 & 90.07\% & 92.88\% & 22.41\% & 37.37\% & 4.33\% & 15.53\% \\
sridutt4598 & 77.67\% & 85.20\% & 22.41\% & 37.37\% & 4.33\% & 15.53\% \\
hharii & 90.07\% & 92.88\% & 10.75\% & 23.35\% & 4.49\% & 15.33\% \\
neelimabhairi & 80.26\% & 86.48\% & 12.85\% & 26.45\% & 6.44\% & 18.13\% \\
javierlorenzo & 53.10\% & 64.10\% & 9.37\% & 21.37\% & -- & -- \\
hari2006 & 90.63\% & 93.84\% & 3.63\% & 13.79\% & 5.14\% & 15.26\% \\
toadmj & 72.41\% & 82.77\% & 4.63\% & 15.69\% & 2.96\% & 14.03\% \\
vivek\_321 & 69.91\% & 80.14\% & 4.29\% & 14.83\% & 5.30\% & 14.99\% \\
vaishu4567 & 27.41\% & 43.14\% & 22.41\% & 37.37\% & 4.33\% & 15.53\% \\
ymlopez & 26.80\% & 45.43\% & 15.08\% & 31.67\% & 3.83\% & 14.81\% \\
sindhura1234 & 5.63\% & 15.87\% & 5.52\% & 16.51\% & 8.44\% & 23.53\% \\
\bottomrule
\end{tabular}%
}
\end{table}

The competition attracted substantial community engagement: 147 participants registered on Codabench, generating 732 submissions during the development phase. For a first benchmark edition on top-view RGB--Depth person Re-ID, this participation level provides a solid empirical basis for comparing methods across tracks.
During the final test window, 33 submissions were recorded on the hidden test subset. This final pool enables a meaningful assessment of competitive performance under a common blind-evaluation protocol. Table~\ref{tab:analysis_aggregate_scores} summarizes aggregate statistics (average and median) for each track.

To complement these aggregate statistics, Table~\ref{tab:participant_full_results} reports participant-level performance across all tracks. All winning-team member names and affiliations are provided in Section~\ref{part:winning_team}.

\begin{table}[t]
\centering
\caption{Aggregate participant statistics per track (average and median over all valid submissions).}
\label{tab:analysis_aggregate_scores}
\resizebox{0.6\textwidth}{!}{%
\begin{tabular}{@{}lccc@{}}
\toprule
\textbf{Track} & \textbf{Metric} & \textbf{Average} & \textbf{Median} \\
\midrule
\multirow{2}{*}{Track 1 (RGB)}
& mAP (Rank-1) & \texttt{89.66\%} & \texttt{93.35\%} \\
& CMC@1 & \texttt{85.73\%} & \texttt{91.92\%} \\
\midrule
\multirow{2}{*}{Track 2 (Depth)}
& mAP (Rank-1) & \texttt{57.19\%} & \texttt{49.82\%} \\
& CMC@1 & \texttt{47.10\%} & \texttt{30.55\%} \\
\midrule
\multirow{2}{*}{Track 3 (RGB$\leftrightarrow$Depth)}
& mAP (Rank-1) & \texttt{40.75\%} & \texttt{23.30\%} \\
& CMC@1 & \texttt{30.39\%} & \texttt{08.79\%} \\
\bottomrule
\end{tabular}
}
\end{table}

\begin{table}[t]
\centering
\caption{Top-3 leaders for each track.}
\label{tab:leaderboard_summary}
\resizebox{0.75\textwidth}{!}{%
\begin{tabular}{@{}llccc@{}}
\toprule
\textbf{Track} & \textbf{Leader} & \textbf{Team} & \textbf{CMC@1} & \textbf{mAP} \\
\midrule
\multirow{3}{*}{RGB Track}
& 1st & \texttt{Hien Pham Duy \etal} & \texttt{100.0\%} & \texttt{100.0\%} \\
& 2nd & \texttt{Carlo Metta} & \texttt{100.0\%} & \texttt{99.98\%} \\
& 3rd & \texttt{Oron Nir \etal} & \texttt{99.84\%} & \texttt{99.92\%} \\
\midrule
\multirow{3}{*}{Depth Track}
& 1st & \texttt{Jin-Hui Jiang \etal} & \texttt{98.86\%} & \texttt{99.38\%} \\
& 2nd & \texttt{Oron Nir \etal} & \texttt{97.19\%} & \texttt{98.34\%} \\
& 3rd & \texttt{Md Rashidunnabi \etal} & \texttt{90.44\%} & \texttt{94.33\%} \\
\midrule
\multirow{3}{*}{Cross-Modal Track}
& 1st & \texttt{Md Rashidunnabi \etal} & \texttt{99.84\%} & \texttt{99.92\%} \\
& 2nd & \texttt{Jin-Hui Jiang \etal} & \texttt{97.89\%} & \texttt{98.76\%} \\
& 3rd & \texttt{Oron Nir \etal} & \texttt{91.50\%} & \texttt{94.88\%} \\
\bottomrule
\end{tabular}
}
\end{table}

\subsection{RGB Track}
Results are shown in Table~\ref{tab:leaderboard_summary}. Here we briefly describe the top three winners of this track.

\paragraph{Hien Pham Duy \etal:}
This team utilized a ViT-Base backbone together with an embedding head producing 512-dimensional descriptors. For training, they combined the official TVRID dataset with two additional top-view person re-id datasets, namely \texttt{TVPR} and \texttt{TVPR2}. Their training pipeline also relied on an identity-aware PK sampling strategy to build batches containing multiple instances of each identity, thereby supporting effective discriminative learning across viewpoints. To improve robustness to the specific challenges of top-view imagery, they applied several human-centered augmentations, including background alternating, background erasing, body-part erasing, local grayscale patch replacement, and global grayscale patch replacement. At inference time, features were extracted from all RGB frames of each tracklet and aggregated by mean pooling to produce a single sequence-level representation. The final retrieval scores were then refined using the k-reciprocal re-ranking strategy proposed by Zhong et al.~\cite{zhong2017re}.

\paragraph{Carlo Metta:}
This team utilized a ConvNeXt-Base backbone\cite{liu2022convnet} for RGB person re-identification, followed by a projection head mapping features to a 512-dimensional embedding space. The training objective combined ArcFace\cite{deng2019arcface} classification loss with batch-hard triplet loss\cite{hermans2017defense}, allowing the model to learn discriminative identity representations adapted to the top-view setting. Their training procedure also relied on identity-aware PK sampling and five-fold identity-disjoint cross-validation, with the best checkpoint retained for each fold. At inference time, embeddings were extracted from up to 48 sampled RGB frames per sequence and averaged to obtain a single sequence-level representation. The final descriptor was further strengthened by ensembling the embeddings produced by the five fold models, and the ranking scores were finally refined using the k-reciprocal re-ranking method of Zhong et al.~\cite{zhong2017re}.

\paragraph{Oron Nir \etal:}
\label{part_rgb:oron_nir}
This team utilized MINER: Modeling Inter- and Intra-Person Variance
for Re-Identification (in-submission for ICML 2026). A contrastive learning framework designed to improve the robustness of person re-identification by explicitly modeling both inter-person and intra-person variance. Built on top of the EVA-CLIP DIFFER\cite{liang2025differ} vision transformer backbone, the method constructs each mini-batch by selecting hard negatives, corresponding to visually similar identities, and hard positives, corresponding to images of the same identity showing strong appearance variations across camera views. Training is then driven by a supervised contrastive objective applied symmetrically between query and gallery samples, encouraging the model to better separate confusing identities while maintaining consistency for the same person across views. The resulting approach therefore focuses on learning view-invariant yet identity-sensitive RGB representations adapted to the top-view re-identification setting.

\subsection{Depth Track}
Results are shown in Table~\ref{tab:leaderboard_summary}. Here we briefly describe the top three winners of this track.

\paragraph{Jin-Hui Jiang \etal:}
\label{part_rgb:jiangvicky}
This team employed a VSLA-CLIP\cite{zhang2024crossplatformvideopersonreid} architecture with a ViT-Base/16 backbone and trained it exclusively on the official TVRID dataset. Their method relied on a two-stage cyclic optimization procedure repeated over five cycles. In the first stage, identity-specific description tokens and shared text prompts were optimized while the encoders remained frozen. In the second stage, the visual encoder was further optimized through trainable temporal adaptation modules and shared learnable prompts injected into the transformer, in order to align visual features with the learned semantic representations. The training objective combined the base VSLA-CLIP loss with additional cross-modal supervised contrastive, cross-modal MSE distillation, and cross-modal InfoNCE losses. Through this cross-modal alignment during training, the learned depth embedding was indirectly shaped by RGB supervision within a shared representation space, which contributed to the strong depth-track performance.

\paragraph{Oron Nir \etal:}
\label{part_rgb:oron_nir_depth}
As in the RGB track (\S\ref{part_rgb:oron_nir}), this team relied on the MINER approach, which fine-tunes the backbone through contrastive learning by jointly emphasizing hard inter-person negatives and hard intra-person positives to obtain a more stable identity-oriented embedding space. For the depth track, the main adaptation concerns input preprocessing: depth maps were converted to inverse depth, normalized with percentile-based clipping, and colorized with a Spectral colormap to produce a pseudo-RGB representation compatible with the backbone. The resulting depth features were learned with the same MINER contrastive formulation as in the RGB case, using camera-view information as weak supervision to encourage view-invariant yet identity-sensitive representations. Final depth rankings were further refined with k-reciprocal re-ranking\cite{zhong2017re}.

\paragraph{Md Rashidunnabi \etal:}
This team proposed a sequence-based depth re-id pipeline in which each passage was represented as a short depth clip composed of sampled frames. Frame-level features were first extracted with a ResNet-50 backbone and then aggregated through temporal attention pooling to produce a single sequence descriptor, further refined by a linear embedding head with batch normalization and L2 normalization. Training relied on paired RGB and depth sequences to learn a shared identity space, while inference in the Depth track used only the depth branch. The optimization combined identity classification, modality-specific batch-hard triplet losses, a cross-modal triplet loss between RGB and depth embeddings, and a center loss to improve intra-class compactness. The training procedure also used PK sampling and clip-based learning over fixed-length sequences, resulting in a temporally aware and cross-modally supervised depth re-identification framework.

\subsection{Cross-Modal Track}
Results are shown in Table~\ref{tab:leaderboard_summary}. Below, we briefly describe the top three teams in this track.

\paragraph{Md Rashidunnabi \etal:}
This team employed XM-VSLA, an adaptation of the VSLA-CLIP\cite{zhang2024crossplatformvideopersonreid} framework for cross-modal RGB--depth person re-identification. Their method used a CLIP ViT-B/16 visual encoder to process fixed-length RGB and depth tracklets within a shared embedding space, while identity-conditioned text prompts were encoded to guide the alignment between visual and semantic representations. Training followed a two-stage procedure: an initial sequence--text alignment stage to establish cross-modal consistency, followed by retrieval-oriented fine-tuning to improve identity discrimination for ranking. At inference time, each sequence was represented by a compact embedding, optionally averaged over multiple clips, and gallery candidates were ranked using cosine distance. The key idea of the approach was to cast RGB--depth re-identification as a vision--language alignment problem in a common embedding space.

\paragraph{Jin-Hui Jiang \etal:}
As in the depth track (\S\ref{part_rgb:jiangvicky}), this team relied on a VSLA-CLIP\cite{zhang2024crossplatformvideopersonreid} framework with a ViT-Base/16 backbone and a two-stage cyclic training strategy. For the cross track, the key difference lies in the objective: the method was explicitly optimized to align RGB and depth samples of the same identity within a shared embedding space through repeated visual-semantic refinement. To this end, training combined the base VSLA-CLIP objective with additional cross-modal supervised contrastive, MSE distillation, and InfoNCE losses, directly reducing the discrepancy between RGB and depth representations. This explicit cross-modal alignment mechanism likely explains its strong performance in cross-modal retrieval.

\paragraph{Oron Nir \etal:}
As in the RGB and depth tracks (\S\ref{part_rgb:oron_nir}, \S\ref{part_rgb:oron_nir_depth}), this team relied on the MINER framework, whose core principle is to structure training around hard inter-person negatives and hard intra-person positives through a symmetric supervised contrastive objective. For the cross track, the key adaptation was to make this contrastive learning explicitly modality-agnostic by directly contrasting RGB and depth representations with a dual-encoder architecture based on the EVA-CLIP DIFFER\cite{liang2025differ} vision transformer backbone (one encoder for RGB and one for depth). In practice, depth inputs were transformed into pseudo-RGB images through inverse-depth conversion, percentile-based normalization, and Spectral colorization, allowing both modalities to be processed in a compatible representation space. This explicit cross-modal contrastive alignment likely explains the strong performance of the approach in RGB--depth retrieval.

\section{Analysis}

\paragraph{Cross-track comparison.}
Following the three official tracks defined in Section~\ref{sec:tracks_description}, Table~\ref{tab:analysis_aggregate_scores} shows a consistent difficulty ordering: Track~1 (RGB) $>$ Track~2 (Depth) $>$ Track~3 (RGB$\leftrightarrow$Depth). 
Relative to RGB, Depth drops by 32.47 points in mAP and 38.63 points in CMC@1; Cross-Modal drops by 48.91 points in mAP and 55.35 points in CMC@1. Between Depth and Cross-Modal, the additional gap is 16.44 points in mAP and 16.72 points in CMC@1. These absolute gaps quantify the expected loss when appearance information is reduced (Depth) or modality-mismatched (RGB$\leftrightarrow$Depth).

At the same time, the top-3 systems in Tracks~2--3 remain strong, indicating that modality-invariant representations are practically achievable on TVRID. This is consistent with recent cross-modal ReID trends (e.g., VSLA-CLIP-style visual-semantic alignment\cite{zhang2024crossplatformvideopersonreid}), where contrastive alignment, prompt-guided optimization, and re-ranking help recover discriminative identity cues even without full RGB appearance.

Based on qualitative inspection of submitted rankings, the most difficult configuration appears to be matching \emph{flat-view} samples against the \emph{oblique} camera. By contrast, the other camera-pair scenarios appear more balanced. Since per-pair quantitative breakdowns are not yet included in the official tables, we keep this observation qualitative.

Track~2 confirms the expected trade-off: depth-only re-identification improves privacy properties while reducing average retrieval accuracy compared with RGB. However, Track~3 also shows that cross-modal matching (RGB query $\rightarrow$ Depth gallery) can be effective, which raises an important question: if depth can be linked reliably to RGB identity signatures, should depth still be considered privacy-preserving \emph{by nature}, or only \emph{conditionally} (depending on threat model, access control, and auxiliary RGB data)? In TVRID, this question is further nuanced by the top-view acquisition setting, where direct face-level appearance cues are limited; therefore, our results support a cautious interpretation: depth is privacy-enhancing, but not automatically privacy-guaranteeing under cross-modal linkage.

\section*{Conclusion}
In this paper, we have presented the ICPR 2026 competition on privacy-preserving top-view RGB--Depth person re-identification (TVRID) and provided a complete report of its benchmark design, protocol, and outcomes. TVRID provides a focused and reproducible evaluation setting featuring 86 identities, four synchronized overhead cameras, structured geometric variation (flat, ascent, descent, oblique), and paired RGB/Depth streams with IN/OUT observations.

The competition protocol was organized around three complementary tracks---RGB Re-ID, Depth Re-ID, and RGB$\leftrightarrow$Depth cross-modal Re-ID---to jointly assess recognition performance, modality robustness, and privacy-aligned deployment scenarios. The final analysis establishes a clear difficulty hierarchy across tracks and quantifies absolute performance gaps under reduced or mismatched appearance cues, while showing that strong methods can still achieve competitive results in constrained modalities through modality-invariant learning.

Beyond leaderboard ranking, the benchmark highlights practically relevant insights: camera geometry remains a key stress factor, depth introduces a measurable privacy--performance trade-off, and effective cross-modal retrieval motivates a more nuanced view of privacy preservation as a conditional property tied to threat models and auxiliary information. In this sense, TVRID serves both as an evaluation benchmark and as a framework for discussing operational privacy risks in modern Re-ID systems.

\section*{Winning Teams and Affiliations}
\label{part:winning_team}
For transparency and proper attribution, the official winners teams in each track are listed below with their member identities and institutional affiliations.

\begin{enumerate}
    \item \textit{Hien Pham Duy \etal}
    \\
    Members: Hien Pham Duy$^1$$^2$, Bao Tran$^1$$^2$
    \\
    Affiliations: $^1$University of Information Technology, VNU-HCM, Vietnam; Vietnam National University, $^2$Ho Chi Minh City, Vietnam

    \item \textit{Carlo Metta}
    \\
    Member: Carlo Metta$^1$
    \\
    Affiliations: $^1$IST CNR, Pisa, Italy

    \item \textit{Oron Nir \etal}
    \\
    Member: Oron Nir$^1$$^2$, Eliyahu Strugo$^2$, Ariel Shamir$^1$
    \\
    Affiliations: $^1$Reichman University; $^2$Microsoft Corporation
    
    \item \textit{Jin-Hui Jiang \etal}
    \\
    Members: Jin-Hui Jiang$^1$, Yu-Fan Lin$^3$, Fu-En Yang$^4$, Yu-Chiang Frank Wang$^4$, Chih-Chung Hsu$^2$$^3$
    \\
    Affiliations: $^1$Institute of Computational Intelligence, National Yang Ming Chiao Tung University; $^2$Institute of Intelligent Systems, National Yang Ming Chiao Tung University; $^3$Institute of Data Science, National Cheng Kung University; $^4$NVIDIA, Taipei, Taiwan

    \item \textit{Md Rashidunnabi \etal}
    \\
    Members: Md Rashidunnabi$^1$$^2$$^4$, Kailash A. Hambarde$^1$, João C. Neves$^2$$^3$, Vasco Lopes$^2$$^3$$^4$, Hugo Proença$^1$$^2$
    \\
    Affiliations: $^1$IT: Instituto de Telecomunicações, Covilhã, Portugal; $^2$University of Beira Interior, Covilhã, Portugal, $^3$NOVA LINCS, University of Beira Interior, Covilhã, Portugal, $^4$DeepNeuronic, Covilhã, Portugal
\end{enumerate}

\section*{Acknowledgment}
This material is based upon work supported by the ANRT (Association nationale de la recherche et de la technologie) in France with a CIFRE fellowship granted to \href{http://www.explainconsultancy.com/en/}{Explain}.

%
%
%
\bibliographystyle{splncs04}
\bibliography{references}

\end{document}